# The Future of Prosody: It's about Time

*Dafydd Gibbon*

**Bielefeld University, Bielefeld, Germany and Jinan University, Guangzhou, China**
gibbon@uni-bielefeld.de

## Abstract

Prosody is usually defined in terms of the three distinct but interacting domains of *pitch*, *intensity* and *duration* patterning, or, more generally, as phonological and phonetic properties of 'suprasegmentals', speech segments which are larger than consonants and vowels. Rather than taking this approach, the concept of *multiple time domains* for prosody processing is taken up, and methods of time domain analysis are discussed: annotation mining with duration dispersion measures, time tree induction, oscillator models in phonology and phonetics, and finally the use of the Amplitude Envelope Modulation Spectrum (AEMS). While frequency demodulation (in the form of pitch tracking) is a central issue in prosodic analysis, in the present context it is amplitude envelope demodulation and frequency zones in the long time-domain spectra of the demodulated envelope which are focused. A generalised view is taken of oscillation as iteration in abstract prosodic models and as modulation and demodulation of a variety of rhythms in different frequency zones.

## 1 Speech and time

### 1.1 Time concepts

In a remarkable *Gedankenexperiment* the philosopher Peter Strawson [1] asked a question which could be taken as part of a phonetician's task list: Can one conceive of an individual, i.e. an object or a person, in a spaceless world with a time dimension only? His answer is a tentative yes, on the assumption that sounds are understood to have dynamically moving sources, interpretable as individual entities. Time, in this sense, is a focal concept in many disciplines, yet in our own research worlds we often act 'as if' temporal segments of speech such as consonants and vowels, syllables, words and other segment sizes were 'objects' rather than dynamic events in time and as if patterns of features were themselves 'objects' rather than parameter trajectories through time. This is ontologically odd, of course, but a useful abstraction.

The present keynote address focuses on domains of processing time and follows various exploratory research and modelling strategies, introducing pointers towards innovative directions of future speech research in different time domains. The standard definition of prosody as *pitch*, *loudness* and *duration* patterning, as 'suprasegmentals' in relation to 'segmental' speech sounds, is supplemented by a rich conception of time which opens up new research perspectives.

In his 1992 ESSLI summer school lecture, Andras Kornai distinguished between two concepts of time: (1) a phonetic concept, which he metaphorically but appropriately termed *clock time*, and (2) *rubber time*, which characterises phonological relations, for example of precedence, hierarchy and inter-tier association in autosegmental and event phonologies, and stress bars in metrical phonologies: it does not matter for a phonological description whether a syllable is 200 milliseconds or 200 minutes long, or, except for practical reasons, 200 years. Clock time is not immediately relevant to the questions asked in phonology, morphology and syntax.

It is convenient to distinguish four time concepts for language and speech, with all four linked by an interpretative or causal chain [2], [3]: *categorial time* (as in a paradigmatic feature such as [± long], *rubber time*, as in *strong* and *weak* properties of nodes in a metrical tree signifying syntagmatic alternation relations between sibling nodes, *clock time*, as a sequence of points and intervals measured with reference to a calibrated clock, and *cloud time*, the everyday analog concept of time, and an essential concept in perceptual phonetics.

The terrain to be traversed in the present study covers the physical-structural divide between clock time and the linguistic domains of rubber time and categorial time, including a generalised view of oscillation as iteration in abstract prosodic models and as rhythmic modulation and demodulation in several frequency zones in the speech signal.

### 1.2 Multiple time domains

There are multiple scales of time dynamics which are addressed in different approaches to language and speech which go far beyond the short processing time domains of the prosodic events we study, and constitute five major time domains, each with its minor subdomains, and in general with fuzzy boundaries between the domains:

1. *discourse time*: the very short time spans, from sub-phone domains to the domains of morphemes, words, sentences, utterances and discourses;
2. *individual time*: years and decades of the acquisition/learning of first and further languages;
3. *social time*: decades and centuries of socially conditioned modifications of language varieties;
4. *historical time*: centuries and millennia of transitions from one state to a typologically distinct state of a language;
5. *evolutionary time*: the multimillenia of language evolution, from emotive prosodies to complex patterns influenced by cultural skills such as rehearsed speech and writing.

These five time domains (cf. also the three time scales of [4]) are all relevant to prosodic studies. Most work in prosody treats part of the first time domain: the trajectories of speech properties through time, of the order of seconds, in some cases minutes, rarely more. Hesitation phenomena and other disfluencies demonstrate the spontaneous dynamic handling of processing constraints in this shortest domain.

The second major time domain, the acquisition of language and speech, sometimes also occupies a very short time span, as with acquisition of individual vocabulary items and idioms, with their prosodic constraints. The domain is receiving an increasing amount of attention, especially in terms of computational learnability models [5], [6], [7].

Prosody in the third major time domain, a fairly popular topic in sociolinguistics and discourse analysis, is shaped by 'influencers' whose prosody is a model for others.

In the fourth major time domain, a familiar case is *tonogenesis*, the emergence of tones, for example from consonantal pitch perturbations [8], and the whimsically named *tonoexodus* ascribed to Matisoff by Henderson [9]: the loss of lexical and morphological tone in the course of time. Perhaps the most famous case of prosody in the fourth time domain is *Verner's Law* (see [10] for a formal account), which describes apparent anomalies in the Germanic consonant systems in terms of accent placement, rather than consonant and vowel contexts. In the fourth time domain, aspects of prosodic form and timing such as intonation are unavailable to direct scientific investigation, although comparative and typological studies may provide hints.

Finally, in the fifth major time domain, some recent studies on language evolution start from simple linear, then linearly iterative grammatical constructions [11]. The prosodic perspective starts much earlier than this [12], with the rhythms and melodies of animal cries. Intuitively, emotional and teleglossic communication both in primate communication and the extreme varieties of emotional human communication, from Lombard speech to shouting, yelling, 'hollerin', howling, and screaming, would form a good starting point, but have so far not received very much attention in studies of prosody.

### 1.3 Processing time

The present approach is concerned with a computationally central dimension of time which is addressed more directly in computational linguistics, phonetics, psycholinguistics, clinical linguistics and speech technology than in linguistics: *processing time*.

One property of speech processing time is central: by default it is linear, meaning that processing time is a linear function of the length of the input and requires only finite memory. That is, patterns are at most left or right branching, with the same processing properties as finite state systems, to which prosodic patterns are easily adapted. The processing time for some patterns, such as centre-embedded nesting, is not linear and in principle requires unlimited memory, and readjustment rules have been proposed for handling the mapping to linear prosody in such models. But it is well-known that human speakers fail when attempting to process more complex varieties of recursion such as centre-embedding and tend to avoid them in spontaneous speech. Attempts to use these constructions are likely to break down beyond a depth of two; cf. [13], [14].

It is not unlikely, speculating a little, that on the time-scale of cultural evolution, centre-embedding arose with the development of writing. Writing provides an additional memory resource, permitting generalisations from unmarked, prosodically straightforward, sentence-final right branching nominal constructions to use of the same structures in highly marked, prosodically problematic and failure-prone centre-embedding, in sentence-initial and sentence-medial positions.

The importance of processing complexity for prosody is often neglected, with definitions left vague or incomplete, and the crucial distinction between linear (left and right branching, finite state) and nonlinear (centre-embedding) recursion not being explicitly made, and the very different processing requirements of speech and writing not being considered [15]. Finite-depth, finite breadth hierarchies (e.g. syllables or simple sentences) need only constant time, or linear time when they iterate. Centre-embedding is fundamentally different. There are experimental studies of centre-embedded read-aloud bracketings, but extensive empirical work on recursion types in spontaneous speech is lacking.

Some theories of grammar propose a finite hierarchy of ranks, from discourse and utterance through sentence and phrase to word, morpheme and phoneme, as in Tagmemics [16], Systemic Grammar (for intonation cf. [17]), Multilinear Grammar and spoken language architecture [2], [13], and, similarly, various versions of the Phonological or Prosodic Hierarchy [18]. Finite depth hierarchies do not add to overall complexity, though the ranks in the hierarchy have their own processing properties, which may add to the complexity.

### 1.4 Outline

The sections following this brief characterisation of the major time domains are devoted to the more conventional discourse time domain and its subdomains. Section 2 discusses annotation mining, and Section 3 deals with rhythmic time patterns as fuzzy similarity, fuzzy alternation and fuzzy isochrony, and with oscillation and the demodulation and spectral analysis of the varying amplitude of the speech signal, the Amplitude Envelope Modulation Spectrum (AEMS). Section 4 investigates two aspects of the discourse time domain in holistic utterance and discourse patterns with the AEMS, and with reference to the prosody of interjections and their surrogates in restricted discourse registers. Summary, conclusion and outlook are presented in Section 5.

## 2 Time in annotation mining

### 2.1 Annotation mining: clock time and interval labels

The term 'annotation mining' itself is about a decade and a half old, though the practice is much older and emerged with the development of statistical modelling in the 1980s and the need to search large databases and archives. The concept emerged independently in domains such as genomics and economics as well as in phonetics [19].

The idea behind speech annotation is the pairing of clock time with transcriptions and other analytic and interpretative information. Annotation mining is information extraction from annotations. From the beginnings in lab-specific software, generic tools which support annotation mining emerged, such as Praat [20], Transcriber [21] and WaveSurfer [22], as well as video annotation tools, making the techniques generally available to linguists, phoneticians, musicologists and media archivists. A general account of annotation is given by the Annotation Graph model [23]. The theoretical foundations of this model are formulated in Event Phonology [24].

### 2.2 Dispersion metrics and timing isochrony

A time domain feature which immediately lends itself to investigation using annotation mining is *rhythm*: consonant-vowel alternation in mora and syllable timing, strong-weak syllable alternation in larger rhythm units. In theoretical phonology there is extensive discussion of rhythm as different kinds of abstract alternation of strong and weak units in relation to words and phrases. The ur-generative and metrical phonologies postulate hierarchies but are agnostic with regard to the 'arity' and the clock time of rhythm patterns in English, while later approaches distinguish, for example, between languages with strict binary and ternary rhythm arity [25], as well as those with more flexible patterns, like English [26].

In phonetic studies during the past few decades, three main approaches have emerged: detailed models of rhythm and duration such as [27], and two which will be discussed further:

the duration dispersion paradigm, and an oscillator coupling and entrainment paradigm. While phonologies concentrate on the alternation relation between consecutive strong and weak rhythm components and are not concerned with issues of timing, the duration dispersion paradigm uses annotation mining and concentrates on timing, but at the expense of the alternation component. In the coupled oscillator paradigm, both timing and alternation are addressed, generally from the production emulation perspective. In the later sections of the present study, the demodulation of oscillations is described, adding a perception emulation perspective.

Several metrics for analysing temporal interval sequences in annotations have been based on the dispersion of phonetic chunk lengths around the mean, including *Pairwise Foot Deviation* (*PFD*) [28], *Pairwise Irregularity Measure* (*PIM*) [29], and raw and normalised versions of the *Pairwise Variability Index* [30]:

$$\text{Variance}(x_{1...n}) = \frac{\sum_{i}^{n}(x_i - \bar{x})^2}{n-1}$$

$$PIM(x_{1...n}) = \sum_{i \neq j} \left| \log \frac{I_i}{I_j} \right|$$
where $I_{i,j}$ are intervals in a given sequence

$$PFD(d_{1...n}) = \frac{\sum_{i=1}^{n} |d - d_i|}{\sum_{j=1}^{n} d_j} \times 100$$
where $d$ is typically the duration of a *foot*

$$nPVI(d_{1...n}) = \frac{\sum_{k=1}^{k-1} \frac{|d_k - d_{k+1}|}{(d_k + d_{k+1})/2}}{n-1} \times 100$$
$d$ refers to duration of vocalic segment, syllable or foot, typically

The dispersion metrics are structurally similar and involve duration averages; they all originated in applications for sets rather than sequences and are isochrony metrics rather than rhythm models: rhythmic alternation is factored out in each case by ignoring the directionality of the duration difference between intervals, either by squaring (*Variance*) or by taking the absolute values of the operations (*PIM*, *PFD*, *PVI*).

The *Variance*, *PIM* and *PFD* metrics apply globally to whole sequences. Their granularity is thus too coarse to account for timing relations between neighbours, and does not take local accelerations and decelerations into account. The *PVI* metrics solve these problems, but at the expense of ambiguity (e.g. *PVI*(2,4,2,4,2,4) = *PVI*(2,4,8,16,32,64) = *PVI*(4,2,1,2,4,8) = 67) and of the exclusion of non-binary patterns: "[the *PVI*] assumes a sufficient predominance of strong-weak alternation in natural usage that the cumulative effect will be to raise the PVI value in a language impressionistically described as stress-timed" ([31], p. 3). This implies that the dispersion metrics are successful heuristic approximations for distinguishing timing patterns, but they beg the question of what rhythm is, and filter out non-binary rhythms as noise. Neither all nor only empirically observable rhythms can be accounted for (*recall*: non-binary patterns excluded; *precision*: non-alternating sequences included).

## 2.3 Annotation mining for time trees

A logical next step is to work on the basic intuition that rhythms are more complex than the two or three levels of consonant-vowel, or syllable-syllable, or syllable-foot relations and retain the alternation property of positive or negative duration differences. The non-alternation problem was solved by Wagner [32] by arranging neighbouring units in a two-dimensional space of z-score normalised pairwise interval durations divided into quadrants for different patterns: long-long, short-short, long-short and short-long.

Figure 1 shows the quadrants for the four binary duration relations between adjacent syllables in English and Beijing Mandarin. The differences between the two timing systems are clear and a quadrant index could be proposed: the ratio of sample counts in strong-strong and weak-weak quadrants. Mandarin syllable durations are scattered fairly evenly around the mean (zero on the z-score scale), while English syllable durations have many short-short timing relations (lower left) and fewer long-long syllable components (upper right), showing a majority of non-binary patterns.

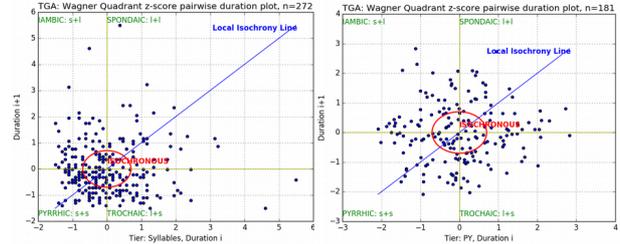

*Figure 1: Wagner quadrant representation of the four binary duration relations between adjacent syllables: English (left), Mandarin (right).*

A further step is to use the difference relations between neighbouring units to induce a hierarchy of long-short relations, a time tree (Figure 2, Figure 3). The algorithm computes a kind of metrical tree from the linear 'clock time' of measured phonetic data rather than the 'rubber time' of intuited phonological data. The procedure is similar to parsing, and can select between strong-weak (trochaic) or weak-strong (iambic) relations, binary or non-binary grouping. Left-right or right-left and bottom-up or top-down processing schedules are possible, but left-right bottom-up is empirically the plausible option.

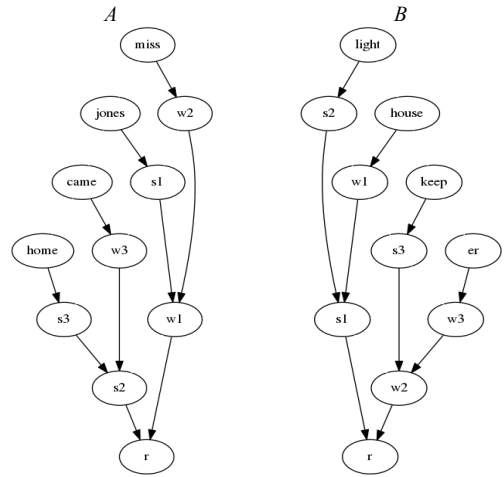

*Figure 2: Tree graphs (root at bottom) induced from numerical sequences: left, iambic (w-s); right, trochaic (s-w) duration relations.*

As an example of how the tree induction algorithm works, a linear sequence of words is paired with duration labels (in this case stylised with faked numbers for the sake of illustration) and metrical tree graphs are induced (Figure 2; tree induction implementations in *Scheme*). Two input-output parametrisations of the algorithm are *iambic* and *trochaic*:

*Iambic* (*weak-strong*) *directionality*:
Input: ((miss . 3) (jones . 2) (came . 3) (home . 1))
Output: (r (w (w miss) (s jones)) (s (w came) (s home)))

*Trochaic* (*strong-weak*) *directionality*:
Input: ((light . 1) (house . 3) (keep . 2) (er . 3))
Output: ((r (s (s light) (w house)) (w (s keep) (w er))))

The inductive procedure, unlike dispersion models, is not based on plain subtraction, but on a 'greater than' or 'smaller than' relation, like the strong-weak relations of generative and metrical phonologies, but the strength of the difference or the ratio can be preserved if required. As an overall algorithm sketch it is sufficient to note the following:

For each interval in a sequence, if $item_i < item_{i+1}$, (if this is the selected relation), join the items into a local tree, store this item and continue until the sequence is finished. Next check the sequence of larger stored items and process in the same way. Repeat this step until no more items have been joined into single tree structures.

The binary parsing schedule can be replaced by an *n*-ary schedule for multiple branching trees, if empirically needed.

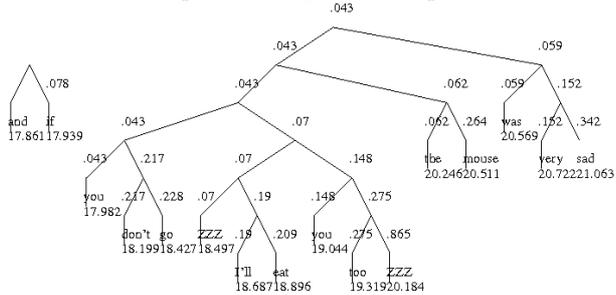

*Figure 3: Time Tree spanning 3 seconds of a story ("The Tiger and the Mouse"), iambic time relations.*

The real interest of this algorithm family emerges when it is applied to long utterance domains such as stories [33]. The application to part of a story ("The Tiger and the Mouse", due to U. Gut) in Figure 3 shows a variety of multiple temporal dependencies or rhythms, from the very long utterance or discourse time domain to highly granular word level relations (the much higher granularity of syllable relations is not reproducible in this print context). Figure 3 shows a time tree induced by annotation mining, showing plausible rhetorical divisions at phrasal and sentence boundaries [26], [27]; quantitative inductive generalisations of this kind do not map deterministically to linguistic categories, of course.

## 3 Alternation, isochrony and oscillation

### 3.1 Basic rhythm model

A sequence perceived as rhythmic is an instance of *oscillation*, and in the ideal case has four main properties [26]: a sequence of *similar* events, *isochrony* (equality of duration among the events), *alternation* (of neighbouring stronger and weaker events) and *iteration* of at least two alternations. Empirical similarity, isochrony and alternation in speech are not mechanically determined but more like *fuzzy similarity*, *fuzzy isochrony* and *fuzzy alternation*, subject to top-down rhetorical, grammatical and lexical factors, and to overriding phonetic timing constraints, such as the increasing duration of strong-weak groups as a function of the number of weak events they contain, for instance unstressed syllables [34]. A generalised view of oscillation is taken here, including iterative abstract models as potential oscillators.

### 3.2 Abstract oscillator models in phonology

The most well-known iterative model of syntagmatic prosodic patterns is Pierrehumbert's finite state grammar of English intonation [35]. The intonation grammar and its derivatives in other studies are usually represented as finite state transition networks, sometimes compactly as a regular expression:

(( %H|%L ( H*|L*|H*+L|H+L*|L*+H|L+H* )+ H-|L- )+ H%|L% )+

The grammar describes groups with choices of boundary tone and pitch accent. The three raised '+' marks indicate potential oscillations: iterable groups with at least one occurrence of the group. The innermost bracketed group contains at least one item selected from the pitch accent lexicon, the next larger bracketed group is a minor or intermediate group of at least one pitch accent sequence, and the outer bracketed group is a major or intonation group containing at least one minor group. The model has been adopted in descriptions of many languages for the description of intonation, with appropriate modifications of the pitch accent lexicon and underlies the popular ToBI transcription conventions.

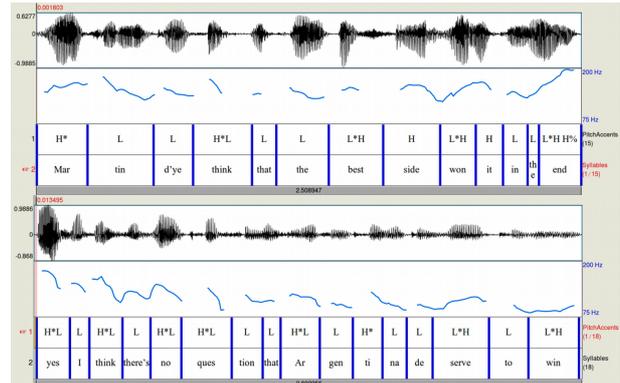

*Figure 4: Question and answer (AixMARSEC J0104G): Praat screenshot showing pitch accent similarity constraints: sequences of variants of H\* then L\* accent types in both question and answer.*

The model as it stands requires extension in three main ways. First, it does not describe the global contour contrasts shown in the globally rising question and falling answer, (Figure 4 and Figure 13, AixMarsec corpus, J0104G [36]). Second, observations of spontaneous speech corpora of English reveal a tendential constraint for iterated pitch accents to be of the same general H or L type (Figure 4). Third, no mapping from phonology to phonetics is defined (but cf. [37]).

Fortunately, a useful feature of finite state models is that they can handle constraints on parallel phonological and phonetic sequences, in 2-tape automata (finite state transducers), enabling a phonetic mapping to be included in the model. An example of phonological-to-phonetic mapping is the generic two-tape oscillator model of [38] and [39], which accounts for terraced tone in Niger-Congo languages. Previously, detailed right-branching tree descriptions [40] had been provided, but without a full grammar.

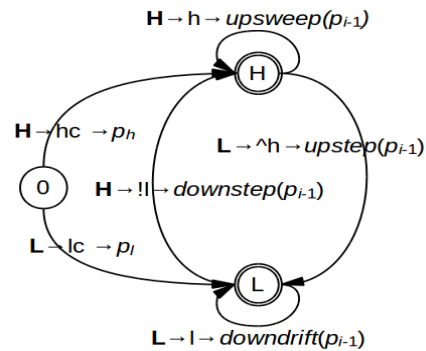

*Figure 5: 3-tape finite state transition network as a tone terracing grammar: start at 0, terminate at H or L.*

Two tapes are not the limit: 3-tape automata can represent two stages of mapping from categorial and rubber time phonological tone representations through categorial phonetic tones to quantitative clock time functions. Figure 5 shows an extension of the 2-tape automaton to a generic 3-tape finite state phonological oscillator model for terraced tone in Niger-Congo languages, formalising the tonal sandhi constraints between pitches of neighbouring tones $p_{i-1}$, $p_i$, starting with quasi-constant high or low pitches $p_h$, $p_l$. The general idea for a given tone is: *assimilation* if the preceding lexical tone is different (upstep, downstep), *reinforcement* of pitch level if the preceding lexical tone is the same (upsweep, downdrift). Semi-terrace loops (at states H and L) and the full-terrace loop (between H and L) define two levels of iteration as potential oscillation. In Figure 5, lexical tone representations are in bold upper case, phonetic representations in lower case ('hc', 'lc' denote initial quasi-constant tones; '!l', '^h' denote phonetic downstep and upstep), *upsweep*, *upstep*, *downdrift*, *downstep* represent quantitative time functions of pitch relative to the preceding phonetic tone.

### 3.3 Modulation and demodulation: oscillator models

A dialectic has emerged between inductive-synthetic models of rhythm as a function of local prominences and holistic-analytic models of emergent rhythm as a function of multiple levels of oscillation in the speech signal, a signal processing paradigm in phonetics which has contrasted with annotation mining approaches since the mid 1990s. Timing models of oscillator coupling and entrainment as a solution to both isochrony and alternation constraints are associated with the pioneering research of O'Dell and Nieminen [41], Cummins and Port [42], Barbosa [43], and have been further developed by the Wagner group [44], [6] and many others. Multiple levels of coupled oscillators, generally pairs of syllable and foot or phrase functions, interact with each other as frequency or amplitude modulation, either within a given speech production, or with interlocutors entraining each other to adapt to the rhythms of their communication partner. In speech the laryngeal carrier signal or noise source is amplitude modulated (effectively: multiplied) by the variable filter of the oral and nasal cavities. Frequency modulation (FM), the basis of the F0 patterns of tones, pitch accents and intonations (and of formant modulations) is not focussed in this paper.

But how can oscillations be dealt with in an emulation of speech perception? One answer is a 'Frequency Zone Hypothesis' of rhythm detection by analysis of the Amplitude Envelope Modulation Spectrum (known variously as AEMS, EMS and AMS) into component frequency zones ('zones' because of the variability of these component frequencies):

1. Demodulate the signal to extract the low frequency (LF) envelope modulation of the waveform.
2. Apply LF spectrum analysis (<50Hz) to the envelope.
3. Examine the low frequency spectrum for oscillations in different, and generally somewhat fuzzy and overlapping frequency zones.

Evidence for multiple rhythms and thus multiple frequency zones in speech was already implicit in much annotation mining: the time-tree induction procedure illustrated in Figure 3, the timing hierarchy of [45], the two levels of [46].

Todd's rhythmogram, based on his Auditory Primal Sketch [47], provides the physiologically most realistic model of rhythm perception in terms of amplitude demodulation and spectral analysis. Perceptually agnostic approaches to rhythm modelling were suggested by [48], [49], [50] and [51] (the last cited as a version of the rhythmogram). The method has previously been used in language identification [52], automatic speech recognition [53], and recently in studies of phonological acquisition [54], [55] and dysarthria [56]. The best known AEMS applications are beat detection in cardiology and music analysis, and for disco lights. The AEMS has been used in a phonetic context, to distinguish the rhythmic properties of typologically different languages [57], resulting in 83.3% correct classification (cf. also [49]), with the conclusion that speech rhythm may be more acoustically encoded than was previously believed.

The AEMS approaches differ in signal processing details:

1. low-pass or band-pass filtering of the signal by median window, Butterworth filter, or, indirectly, down-sampling;
2. amplitude demodulation by
   1. rectification: half-wave 'diode' rectification, positive amplitude, vs. full-wave rectification, absolute amplitude;
   2. envelope extraction by filtering or peak-picking;
   3. alternatively, demodulation by Hilbert transform [58];
3. envelope smoothing by low-pass filtering (see #1);
4. windowing and LF spectral analysis of the envelope using
   1. filter banks covering a range of frequencies,
   2. Fourier transform,
   3. other related techniques;
5. spectrum processing to detect frequency zones.

For the visualisations in the following subsections, a minimal model of full-wave demodulation, envelope extraction by peak-picking, envelope smoothing and Fast Fourier Transform is used, with no windowing. Several of the filtering and windowing signal processing stages used in existing models are bypassed, and the carefully tailored data chunking of several earlier approaches is replaced by raw recordings of read speech with chunks from 5s to 30s, for show-casing long-term utterance and discourse rhythms. A stylised AEMS application is shown in Figure 6 to illustrate the method (AEMS analyses are implemented in *Python*, pitch estimation with *RAPT* [59])

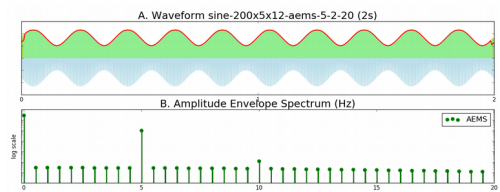

*Figure 6: A, top: Calibration signal, 200 Hz carrier, 5 Hz modulation, rectification and envelope. B, bottom:AEMS.*

A 200Hz synthetic signal (thus, in the female voice range) of 2s duration is amplitude modulated with a 5Hz (200ms) 'rhythm' (thus, in the syllable frequency range). In Figure 6A, the absolute amplitude values (green, in colour rendering) are obtained by removing the negative component (pale blue), and the envelope (top line, red) is extracted. In Figure 6B, the AEMS generated by Fast Fourier Transform is visualised, with the demodulated 5Hz 'rhythm' clearly shown in the spectrum above the digital noise (the weaker 10Hz $2^{nd}$ harmonic is a distortion: the digital signal is not a perfect sinusoid).

### 3.4 AEMS: visual time domain exploration

AEMS techniques are promising rhythm models. The following section explores further avenues within the present framework of methodological exploration, prioritising visualisation of basic principles for directly contrasting rhythmic effects from English (Edinburgh and AixMARSEC corpora) and Mandarin (Yu corpus [60]). The objective of the discussion is exploration, not empirical proof.

The null hypothesis, that AEMS frequencies are smoothly distributed, is not expected. Plausible alternative hypotheses are derivable from the Wagner quadrant and the time tree approaches. A language with word stress, sentence stress, and variable pitch accent realisations of stress, such as English, would be expected to have a different AEMS shape from that of a language with lexical tone and a typologically different grammar, such as Mandarin Chinese. Figure 7 and Figure 8 show the low-pass filtered AEMS with values <5Hz (200ms) for 10s segments of readings of "The North Wind and the Sun" in English and Beijing Mandarin, respectively, with the shape of the AEMS highlighted by superimposition of matching $9^{th}$ degree polynomials (the minimal complexity for visually fitting the spectrum) rather than detailed splines [61] or time function components [62].

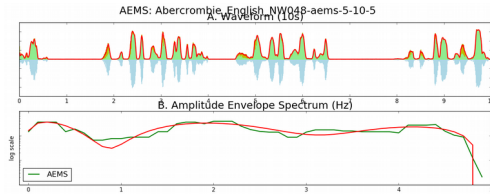

*Figure 7: Male, RP, Edinburgh corpus NW048. Upper: 10s of "The North Wind and the Sun", B, lower: AEMS.*

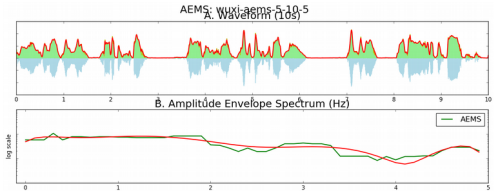

*Figure 8: Beijing Mandarin (young adult female, Yu corpus, wuxi). A, upper: 10s of "The North Wind and the Sun",. B, lower: AEMS.*

In English the two frequency zones around 2Hz (500ms) and 4.25Hz (240ms) are compatible with the prediction of frequency concentrations for stressed syllables and strong stressed syllables. For Mandarin the result shows frequency zones around 4.75Hz (210ms) and 3Hz (33ms) which are compatible with a distinction between tone-bearing syllable and polysyllabic word domains. Small alternations in syllable durations <100ms occur, especially with the neutral 'fifth tone', a possible explanation of this frequency distribution, and a pointer to further research.

A further comparison of the AEMS (<20Hz, >50ms) of 10 Mandarin recordings (Yu corpus) and 5 English recordings (AixMarsec corpus) were found to be different, $t$=12.3, $p$<0.01, in line with the conclusion of [57].

### 3.5 Discourse rhythms

The coupled oscillator approaches account for multiple rhythmic time domains in speech, and [46] investigated multiple time domains with a dispersion metric. The AEMS method is suitable for examining multiple very long term rhythms, the waves and tides of speech rather than the ripples, with discourse patterns and rhetorical rhythms due to pauses, emphasis, intonational paragraphs or 'paratones' [63], [64]. Three segments each of 10s and 30s duration (A1202B in the Aix-MARSEC corpus, wuxi in the Yu corpus), are illustrated in Figure 9 and Figure 10 for frequencies below 1Hz ($\delta t$>1s). A heatmap format is chosen for the AEMS as a kind of 'histogram in colour'.

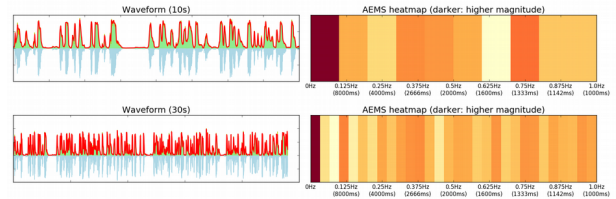

*Figure 9: Heat maps of 10s, 30s waveforms and <1Hz AEMS spectra (A1202B, newsreading, AixMARSEC corpus).*

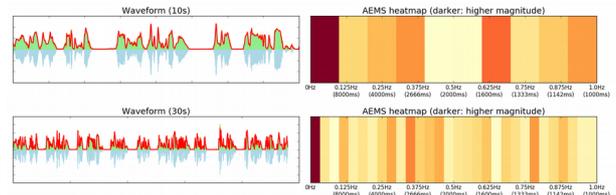

*Figure 10: Heat maps of 10s, 30s waveforms and <1Hz spectra (wuxi, story reading, Yu corpus).*

Figure 11 shows the frequency zones of different discourse prosodic domains in a hierarchical segmentation of the z-scored spectra of Figure 9 (top) and Figure 10 (top), using the same algorithm family as was used for Figure 2 and Figure 3.

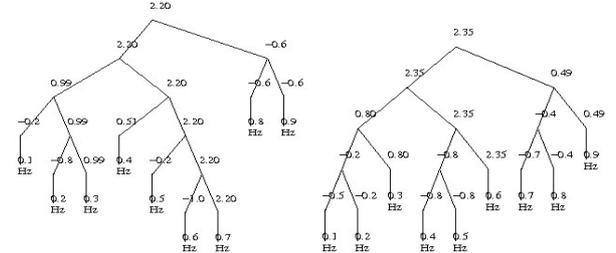

*Figure 11: Spectral hierarchy <1Hz for 10s segments of English (left) and Mandarin (right).*

The English frequency zone at 0.75Hz (1.3s) in the <1Hz rhythm spectrum of Figure 9 (top), appears as a major fork in Figure 11 (left). This frequency zone tends to represent shorter phrases and the 0.35Hz zone pause distributions.

The Mandarin frequency zones in Figure 10 (top) are very differently structured, as Figure 11 shows. The Mandarin rhythm spectrum has a much clearer break around 0.6Hz between the lower frequency pause zone and the higher short phrase frequency zone, and there is more frequency variation.

The frequency zones for the <1Hz spectrum over the longer 30s stretches of speech show more differentiation. To what extent the differences and the variations relate to differences in tempo and genre, individual differences and typological language differences, is a matter for further detailed research. As with the other visualisations, the intention is exploratory. Further quantitative analysis will need to proceed with investigation of different time domains, different analysis windows and different granularities, from very short to very long, with the development of appropriate feature vectors to permit reliable classification and comparison of prosodically different languages, genres and styles at all ranks, of discourse and utterance, as well as the more customary sentence, word and sub-word ranks.

Preliminary tests with spectral analysis of F0 tracks have initially shown similar effects to the AEMS analyses, which is not really a surprising result (cf. also [50]). However, long time domain F0 spectrum analyses need further investigation in order to determine how far long time domain F0 spectra are independent of the AEMS.

## 3.6 Pitch in the discourse time domain

Pitch patterns over long time domains have received much attention with qualitative ethnomethodological and other discourse-analytic methodologies, but there has been little quantitative modelling over longer time domains.

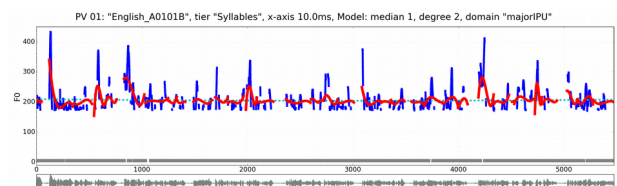

*Figure 12: RP English female newsreader (BBC, AixMARSEC corpus A0101B, 57s): F0 track with superimposed matching 3rd degree polynomial model, time labels in cs. Pitch visualisations in Python.*

Figure 12 shows a sequence of news items read aloud by a female newsreader (AixMARSEC corpus A0101B), illustrating iterative pitch patterning, with high F0 onset followed by a downtrend and final falling F0. A matching 3rd degree polynomial model is superimposed in order to visually smooth the overall tendencies. The monologue discourse divides into paratones, each of which has an extremely high F0 onset in the region of 400Hz, a distinctive characteristic of many female Southern British speakers. The iterative pattern shows a 'flat hierarchy' with a high paratone onset level, and lower, though still high, onset pitch levels for utterances within the paratone. The pattern can be modelled as oscillation by iteration) as in the Pierrehumbert model.

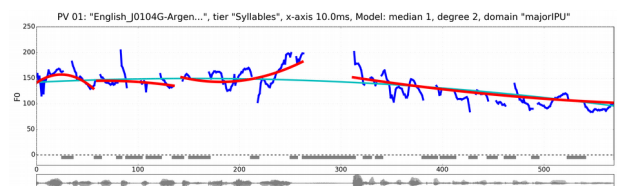

*Figure 13: Interview Question-Answer sequence (2 male adult RP speakers) with superimposed matching 2nd degree polynomial contours over IPUs, 27.7s, time in cs.*

Figure 13 (cf. also Figure 4) shows a question-answer sequence from a broadcast interview (AixMARSEC J0104G). Matching 2nd degree polynomial contours are superimposed in two time domains in order to mark the contour trends (for polynomial fitting of intonation curves cf. [65], [66] and contributions to [67]. Two interpausal Units (IPUs) in the question show an overall falling-rising contour, the answer shows an overall falling contour, and the global contour over both question and answer follows a holistic rise-fall pattern.

Investigation of discourse intonation patterns such as these promise empirically based insights into time domains which go beyond the time domains of word and sentence processing.

## 3.7 'Wow', whistles and the Mandarin '6th tone'

The prosodic categories which have been studied in the discourse time domain are mainly focus and other varieties of stress pattern semantics which have been generalised from the sentential time domain. But there are patterns in the very short time domain which are restricted to the discourse rank and do not combine with lower ranks: the pitch contours of calls (cf. [68], [69]) and interjections. F0 traces of five emotive interjections elicited from a female speaker of Cantonese Mandarin are shown at the top of Figure 14. The rise-fall tones are clearly distinct from the four lexical tones of Mandarin (bottom of Figure 14), and from the neutral 5th tone. This '6th tone' has intonation status, as a discourse tone rather than a lexical tone. The contours are visually highlighted with a matching 2nd degree polynomial contour, cf. [70], [71], [72].

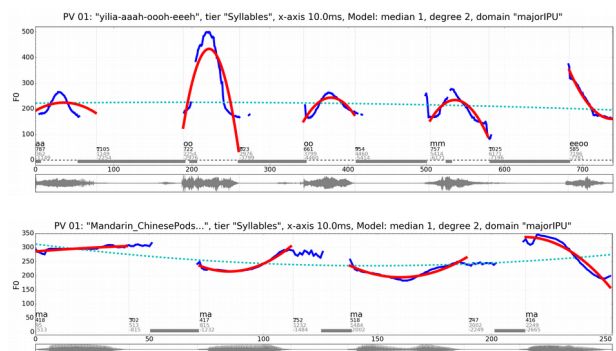

*Figure 14: Upper: Female, Mandarin, elicited monosyllabic interjections with 'Tone 6' (rise-fall) and Tone 4 (fall), first syllable terminates with creak, 2nd syllable includes falsetto phonation; lower: female, Mandarin, elicited syllable "ma" with the 4 Mandarin tones.*

Presumably most languages do not have a highly structured whistling register like those of the mountainous regions of La Gomera, Greece, Turkey, and Mexico [73]. Nevertheless, whistling, particularly emotive street whistling, occurs in many other languages as a surrogate for interjections, calls and exclamations. For example, in English, a rise-plus-rise-fall pitch contour associated with the appraisive exclamation "Oh my!" or "My God!" has the same rise-plus-rise-fall frequency pattern as the 'wolf whistle' and shares the rise-fall component with the second formant of the interjections written as "Whew!" and "Wow!", associated with similar resonance configurations in speech production.

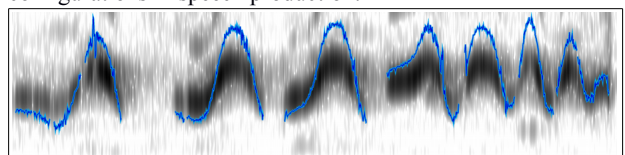

*Figure 15: Cropped Praat graph for a rise-fall street whistle sequence by a Cantonese schoolchild (Guangzhou), whistle range 1-2kHz; the superimposed F0 trace shows 1 octave lower due to aliasing.*

Figure 15 shows the F0 contour (1-2kHz) in a spectrogram of an iterated street whistle by a Cantonese schoolchild. The hypothesis is that, as with the English wolf-whistle, the F0 shape of the whistle contour matches the F0 shape of an interjection in the discourse domain. The F0 of the whistle is too high for regular speech F0 track settings, and the overlaid trace detects a pseudo-subharmonic one octave lower than the actual F0, due to aliasing. Quantitative measures are less important here than documentation of the contour, and the aliased F0 trace highlights the shape of the contour very well, demonstrating the similarity of the rising-falling contour of the whistle and of the rising-falling contour of the interjections (Figure 14), both of which are apparently documented here for the first time. There are strong similarities between primate calls and these whistles (and the tones of the 'wow' and 'aah' exclamations), which suggests that they may have occurred at a very early date in the development of speech prosody in the evolutionary time domain.

## 4 Summary, conclusion and outlook

Domains in linguistics are units to which rules apply, or to constituents of language structures, with no reference to 'clock time', though with implicit reference to 'rubber time' temporal relations. Domains in the 'clock time' of speech analysis are either the time domain or the frequency domain. The present approach takes a syncretistic view of various time domain concepts, and addresses the multiple temporal domains which are constitutive for the field of prosody seen as a whole.

The first major time domain (utterances, discourse) is just as dynamic as the others, in the sense that speech formulation and decoding strategies may change in real time, as with hesitation phenomena, and learning (for example of vocabulary) may occur even within very short sentential time domains. That the major time domains of language acquisition, and social, historical and evolutionary time, involve dynamic change is inherent in their definition. Much is known about short domain speech prosody, and very little is known about the longer major domains.

First, a number of issues in prosodic phonology were discussed, pertaining to limitations on recursion types which underlie the oscillatory principles of 'rubber time' rhythm. Second, a number of 'rubber time' and 'clock time' approaches to speech timing hierarchies were discussed, from duration dispersion metrics through oscillator emulations of rhythm production in speech to an annotation-independent approach to rhythm analysis, the amplitude modulation spectrum analysis of speech, AEMS, and the Frequency Zone Hypothesis of rhythm, interpretable as emulation of perception. The incomplete exploratory analyses used to illustrate these concepts require much further research, for example of the development of the functionality of the AEMS in evolution.

Although it is fascinating to browse through the processing of multiple temporal domains, of course the everyday work of a scientist involves atomisation of a given domain into manageable chunks, involving experimental, observational and hermeneutic analysis, and reconstruction of larger wholes. This keynote contribution is designed to encourage 'thinking outside the box' with unfamiliar strategies, looking at innovative methods which scholars have developed in neighbouring disciplines as shown in this study, and, building on these, developing further innovative methods. The examples in this contribution are chosen more for illustration by exploratory visual inspection and illustration than for quantitative empirical proof, following the example of initial exploration in 'big data' projects, but here with 'tiny data'.

There are many methodological boxes to be opened, re-thought and re-examined with care and new insight. And there are many wide-open domains for the study of prosody and its applications, especially in the longer major time domains. The future of prosody? It's about time.

## 5 Acknowledgments

This study is dedicated to the memory of the late Wiktor Jassem, emeritus of the Polish Academy of Sciences and pioneer in spectral analysis, in gratitude for friendship, mentorship, co-authorship, and for many fruitful discussions of his original work, of the issues addressed in the present context, and of prosody in general. I am indebted very many colleagues for discussions over the years, and specifically to my colleagues Yu Jue, Liang Jie, Lin Xuewei, and my students Li Peng, He Linfang, Feng Baoyin and Bi Dan for data and discussion of interjection tones and street whistles, to Alexandra Gibbon for support with *R*, and to Erwin Jass for advice to implement models physically or computationally:

"If it ain't tested, it don't work."